\def\year{2022}\relax
\documentclass[letterpaper]{article} 
\usepackage{aaai22}  
\usepackage{times}  
\usepackage{helvet}  
\usepackage{courier}  
\usepackage[hyphens]{url}  
\usepackage{graphicx} 
\usepackage{natbib}  
\usepackage{caption} 
\DeclareCaptionStyle{ruled}{labelfont=normalfont,labelsep=colon,strut=off} 
\usepackage{algorithm}
\usepackage{algorithmic}
\usepackage{multirow}
\usepackage{amsmath}
\usepackage{amssymb}

\newlength\savewidth\newcommand\shline{\noalign{\global\savewidth\arrayrulewidth
  \global\arrayrulewidth 1pt}\hline\noalign{\global\arrayrulewidth\savewidth}}
\usepackage{newfloat}
\usepackage{listings}
\floatstyle{ruled}
\newfloat{listing}{tb}{lst}{}
\floatname{listing}{Listing}
\nocopyright

\title{Contrastive Video-Language Segmentation}
\author {
    Chen Liang,\textsuperscript{\rm 1}
    Yawei Luo, \textsuperscript{\rm 1}
    Yu Wu \textsuperscript{\rm 2}
    and Yi Yang \textsuperscript{\rm 1}
}
\affiliations {
    \textsuperscript{\rm 1} Zhejiang University \\
    \textsuperscript{\rm 2} University of Technology Sydney \\
}

\begin{document}

\maketitle

\begin{abstract}
We focus on the problem of segmenting a certain object referred by a natural language sentence in video content, at the core of formulating a pinpoint vision-language relation.
While existing attempts mainly construct such relation in an implicit way, \textit{i.e.}, grid-level multi-modal feature fusion, it has been proven problematic to distinguish semantically similar objects under this paradigm.
In this work, we propose to interwind the visual and linguistic modalities in an explicit way via the contrastive learning objective, which directly aligns the referred object and the language description and separates the unreferred content apart across frames.
Moreover, to remedy for the degradation problem, we present two complementary hard instance mining strategies, \textit{i.e.}, Language-relevant Channel Filter and Relative Hard Instance Construction. They encourage the network to exclude visual-distinguishable feature and to focus on easy-confused objects during the contrastive training. Extensive experiments on two benchmarks, \textit{i.e.}, A2D Sentences and J-HMDB Sentences, quantitatively demonstrate the state-of-the-arts performance of our method and qualitatively show the more accurate distinguishment between semantically similar objects over baselines.
\end{abstract}

\section{Introduction} \label{sec:introduction}
Video-language segmentation (VLS) aims to predict a set of segmentation masks for the referent in a video under the guidance of a natural language description. This practical problem is fundamental to many applications such as interactive video editing, object tracking and human-machine interaction~\cite{wang2021survey}. Unlike the general semantic segmentation tasks that is operated over a fixed set of pre-defined categories, VLS requires fine-grained joint modeling for vision and language, yielding a more challenging scenario due to the great diversity of linguistic concepts and extra semantics in temporal domain.

\begin{figure}[t]
\begin{center}
     \includegraphics[width=0.98\linewidth]{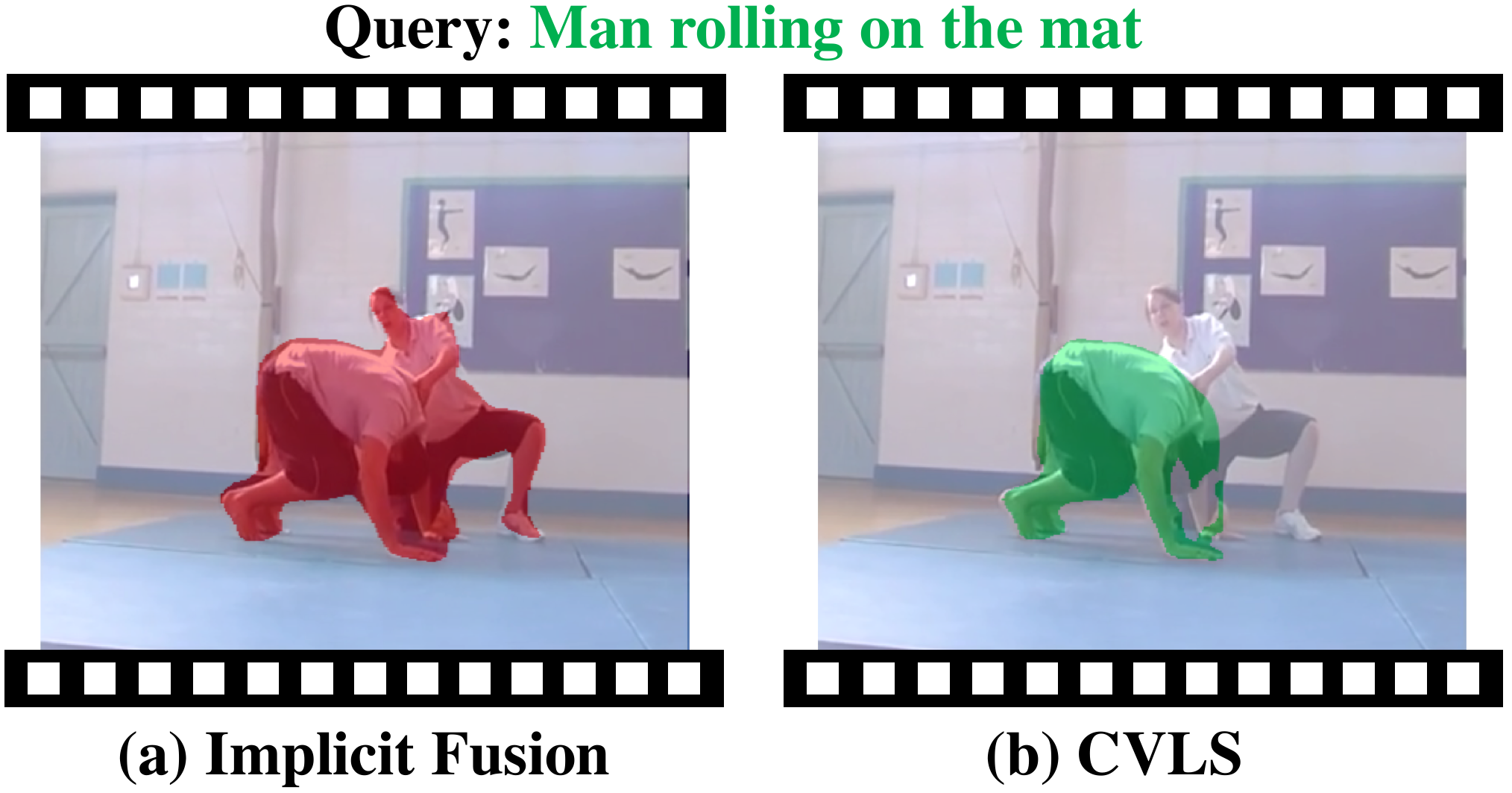}
\end{center}
\captionsetup{font=small}
\caption{\small
      \textbf{Illustration of the video-language segmentation task.}
      Given a linguistic expression, the predictions of state-of-the-art method \cite{wang2019asymmetric} and CVLS are shown in (a) and (b) respectively.
      Previous methods (a) mainly rely on the implicit fusion to formulate cross-modal interaction and fail to distinguish semantically similar objects while CVLS (b) does the opposite.
      }\label{fig:motivation}
\end{figure}

A commonplace solution in prior work is first extracting visual and linguistic features respectively, followed by a fusion process to implicitly interwind cross-modal features, \textit{e.g.}, straightforward concatenation \cite{shi2018key,hu2016segmentation}, cross-modal attention-mechanism \cite{wang2019asymmetric,ningpolar,hui2021collaborative} or dynamic filter \cite{wang2020context,gavrilyuk2018actor}, and then conducting segmentation based on the fused feature. However, these solutions rarely consider the crucial fine-grained alignment of the cross-modal information hence prone to fail in distinguishing the semantically similar objects. As  in Fig.~\ref{fig:motivation}, the model with implicit fusion strategy fails to capture the contextual difference among multiple similar regions, \textit{i.e.}. person conducting different actions.

In this paper, we focus on explicitly modeling the vision-language interactions for the VLS problem. Our approach is inspired by the contrastive representation learning theory~\cite{wu2018unsupervised,chen2020simple,wang2021exploring} whose fundamental purpose is to capture similarities between examples of the same class and contrast them with examples from other classes. Dubbed Contrastive Video-Language Segmentation (CVLS), our method presents a cross-modal contrastive objective to align the referred visual object with the linguistic description in representation space while contracting the unreferred video information with the description at the same time. In this way, fine-grained vision-language interactions can be explicitly learned thus boosting the representation discriminability among visually similar objects based on a linguistic description.

Although vanilla contrastive representation learning provides a potentially feasible solution for the vision-language interactions, we find it problematic to apply the Na\"ive paradigm mechanically on the raw cross-modal representations.
Upon reviewing the VLS task, we notice that the background areas occupy a large proportion across frames while the visually similar candidates do the opposite.
Since the background areas can be easily excluded from the candidate referent just using the visual features while the visually similar objects can be only distinguished on condition of a linguistic feature\footnote{For convenience, we use ``visual-distinguishable region'' to refer the former while use ``linguistic-distinguishable object'' to refer the latter hereafter.}.
For example, as shown in Fig.~\ref{fig:motivation}, models can easily distinguish foreground objects (person) apart from a large area of background even without a linguistic guidance.
A trivial construction of instance-level positive-negative pairs would introduce excessive uninformative visual samples that are over-easy for the current linguistic embedding to discriminate, which might directly degrade the cross-modal contrastive objective to a single-modal feature classifier, \textit{i.e.}, a visual saliency detector.

To address this issue, we present two simple yet effective hard instance mining strategies, \textit{i.e.}, Language-relevant Channel Filter (LCF) and Relative Hard Instance Construction (RHIC), which are specially designed for the cross-modal scenario.
They encourage the network to exclude uninformative visual-distinguishable features and to focus on linguistic-distinguishable objects during the contrastive training.
On one hand, with the guidance of linguistic description, LCF module generates channel-wise filters for visual features to highlight the linguistic-distinguishable features.
On the other hand, in RHIC module, only those misclassified pixels with higher confidence are considered while positive-negative pair sampling, to avoid gradient from being dominated by samples from visual-distinguishable regions.
Main contributions of our work are as follows:
\begin{itemize}
\item We propose CVLS, a cross-modal contrastive learning-based method for VLS task, which explicitly captures the commondality of vision-language representations thus boosting representation discriminability among visually similar objects on condition of linguistic descriptions.
\item We propose two simple yet efective hard mining strategies tailored for cross-modal contrastive objective. These strategies are able to filter out over-easy video content from visual features, which enriches the informative gains for contrastive learning.
\item The proposed method significantly outperforms SOTA methods in most metrics on two popular video-language segmentation benchmarks, \textit{i.e.}, A2D, and J-HMDB.
\end{itemize}

\section{Related Work}
\subsection{Video-Language Segmentation}
Towards understanding fine-grained video representation, Xu \textit{et al.}~\cite{xu2015can} introduce a task of actor and action video segmentation with the Actor-Action Dataset (A2D) containing a fixed set of actor-action pairs and pixel-level annotations.
Later,
Gavrilyuk \textit{et al.}~\cite{gavrilyuk2018actor} extend Actor-Action Dataset (A2D) with human-annotated sentences and introduce the challenging text-based video segmentation task that requires further comprehension of both vision and language modalities.
They adopt dynamic convolution filters generated by language to align the multi-modal feature.
Then, Wang \textit{et al.}~\cite{wang2020context} extend vanilla dynamic convolution with vision context to generate spatial-relevant kernels. Ning \textit{et al.}~\cite{ningpolar} embed spatial expressions into attention module in terms of direction and range for better linguistic spatial formulation. Wang \textit{et al.}~\cite{wang2019asymmetric} utilize asymmetric attention mechanisms to facilitate visual guided linguistic feature learning. Hui \textit{et al.}~\cite{hui2021collaborative} further introduce an additional 2D spatial encoder to alleviate the spatial information misaligned problem brought by 3D CNNs.
Mcintosh \textit{et al.}~\cite{mcintosh2020visual} introduce a capsule-based approach for better capturing the relationship between multi-modal features. To further mining continuous temporal information, they extend the A2D dataset with annotations for all frames.
Even with certain achievement, existing methods mainly suffers from coarse cross-modal interaction, \textit{i.e.}, implicit multi-modal fusion, which directly leads to ambiguous boundaries for semantically similar objects.
In this paper, we address this limitation by constructing instance-level multi-modal interaction in an explicitly contrastive manner.

\subsection{Multi-modal Contrastive Learning}
Contrastive Learning techniques learn representations in a discriminative manner by pulling positive pairs closer against negative pairs.
Recently, self-supervised visual representation learning surges and earns a great success either by partial data prediction \cite{liu2020pic,doersch2015unsupervised,dosovitskiy2015discriminative}, or instance-level training image discrimination \cite{chen2020simple,he2020momentum,tian2019contrastive,wu2018unsupervised}. These methods mainly rely on instance augmentation to form positive pairs.
Beyond augmentation-based approach,  multi-modal contrastive learning leverages multiple modalities of the same instance to construct positive pairs. And cross-modal representations are learned during the procedure.
Examples of commonly used data modalities include appearance of image, motion of video \cite{harley2019learning}, depth \cite{tian2019contrastive}, luminance and chrominance \cite{zhang2017split}, audio \cite{wu2019DAM,wu2021explore}, and text \cite{zhang2020contrastive,gupta2020contrastive} as in our work.
Zhang \textit{et al.}~\cite{zhang2020contrastive} and Gupta \textit{et al.}~\cite{gupta2020contrastive} both design sophisticated strategies to align vision region with expression and maximize the mutual information between the modalities. In this work, inspired by such great progress, we propose a specially-designed cross-modal contrastive learning objective for the first attempt of introducing explicit cross-modal interaction into Video-Language Segmentation.

\begin{figure*}[t]
\begin{center}
     \includegraphics[width=0.9\linewidth]{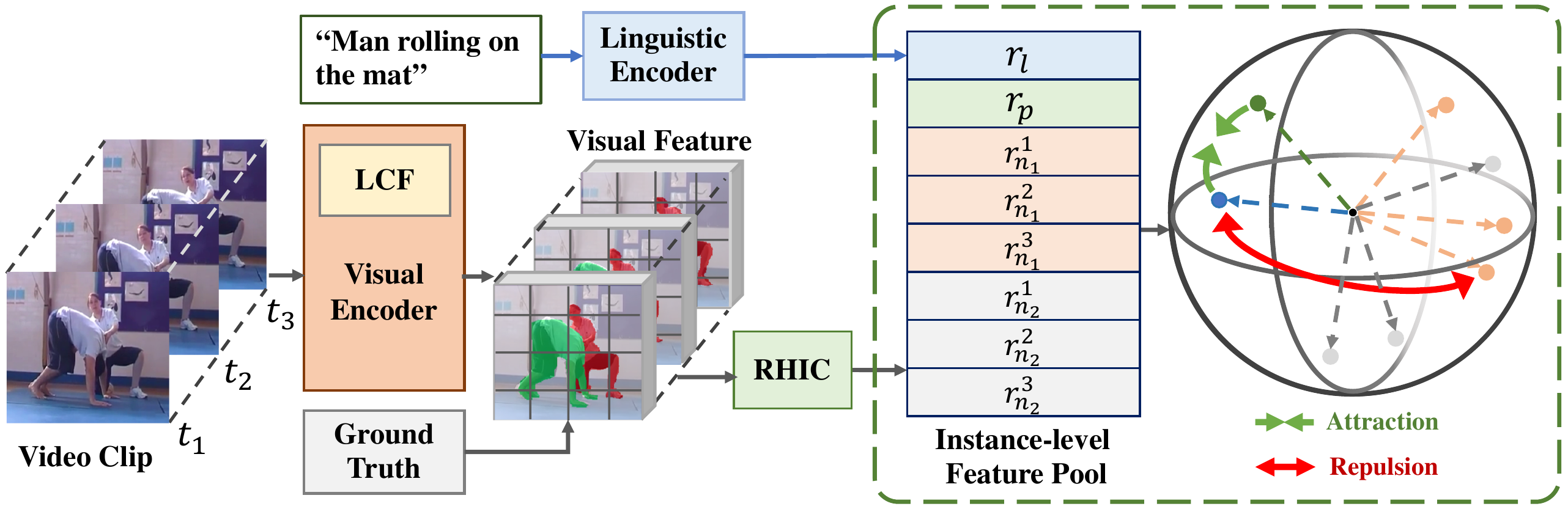}
\end{center}
\captionsetup{font=small}
\caption{\small
\textbf{Illustration of Contrastive Cross-modal Learning Module (CCLM)}. Visual encoder and linguistic encoder are applied to extract object-level visual features $\{r_p, r_{n_1}^1, \cdots, r_{n_2}^3\}$ and reference feature $r_l$ respectively, which are then projected to a $128$-dimensional space and $L2$ normalized.
$r_p$ is the corresponding referent feature.
$r_{n_i}^t$ denotes negative instances in the video with time step $t$, including background (grey region).
Optimal representations are learned via cross-modal discrimination, where the referred object and the language description are pulled closer and the distracting contents are pushed apart across frames.
Language-relevant Channel Filter and Relative Hard Instance Construction are abbreviated as ``LCF" and ``RHIC" (Sec.~\ref{section:LCF}) respectively.
}\label{fig:pipeline}
\end{figure*}

\section{Method}
In this section, we first present the basic Cross-modal Conrastive Learning procedure in Sec.~\ref{section:ObjCons}. To further alleviate degradation problem, we then introduce two specially-designed cross-modal hard mining approaches in Sec.~\ref{section:HM}. Even not being restricted to any certain feature extractor, in Sec.~\ref{section:overall}, we finally describe the overall architecture implemented in this work.

\subsection{Na\"ive Instance-level Cross-modal Contrastive Representation Learning (CCLM)} \label{section:ObjCons}
%
As illustrated in Fig.~\ref{fig:pipeline}, with a pair of reference (natural language description) and corresponding referent (visual object), our goal is to learn a good representation space that the positive pairs would be pulled together while other objects in a video would be pushed away at mean time.
\subsubsection{Within-modal representation construction.}
Different from the prevailing segmentation models that formulate visual embeddings in grid-level as in \cite{zhao2020contrastive}, in VLS task we argue that it is more reasonable and efficient to formulate within-modal embeddings and construct cross-modal commonalities in instance-level concerning that natural language expressions are mostly annotated in an object view.
Now the problem lies in how to construct within-modal representations for the reference and objects.
For linguistic modality, reference feature $r_l$ can be directly extracted by a sentence-level linguistic encoder (Detailed later in Sec.~\ref{section:LE}).
For visual modality, instance-level feature is obtained by conducting average-pooling on mask-cropped visual feature map extracted from visual encoder (Sec.~\ref{section:VE}).
Formally, the naive process for obtaining instance representation $r_v^t$ at time step $t$ could be defined as:
\begin{equation}
r_v^t = AVG(X^t_v \odot o^t). \label{equation:OT}
\end{equation}
Here we denote $X^t_v$ and $o^t$ as the extracted visual feature and a binary object mask.
$\odot$ and $AVG$ correspond to element-wise multiplication and average pooling respectively.

We treat the entire video clip as a pool of object features $\{r_{v_1}^1,r_{v_1}^2,\cdots,r_{v_{N_v}}^T,\}$, where $N_v$ is the total number of instances in a video clip\footnote{Following definitions in literature~\cite{lin2021video,wang2021end}, we use the term \textit{instance} for video-level identification and use \textit{object} for frame-level, \textit{e.g.}, there are $2$ instances and $2T$ objects in the example video clip (Shown in Fig.~\ref{fig:pipeline})} with $T$ frames.
For ease of notation, we define the positive referent object corresponding to each linguistic reference $r_l$ as $r_p$ and all other negative visual objects as $r_{n_i}^t$, where $i \in \{1,\cdots,N_v-1\}$ and $t \in \{1,\cdots,T\}$.
Correspondingly, $o_p$ is the referent region and $o_{n_i}^t$ is the negative object region. Both of them can be obtained from ground-truth masks during training. In later sections, we omit time stamp $t$ for ease of notation.
%
It should be noted that, background region (grey region in Fig.~\ref{fig:pipeline}) is also considered as a negative instance and counted in $N_v$.

Within-modal features in the pool are further fed into a projection network, $Proj(\cdot)$, mapped to a same cross-modal representation space as in Fig.~\ref{fig:pipeline}, \textit{i.e.}, $z_p = Proj(r_p)$, $z_{n_i} = Proj(r_{n_i})$, $z_l = Proj(r_l)$.
We instantiate $Proj(\cdot)$ as a multi-layer perceptron as in~\cite{chen2020simple} with a single hidden layer of size $512$ and output vector of size $128$.
The outputs are then normalized to lie on an unit hypersphere, which enables using an inner product to measure distances in the projection space, \textit{i.e.}, $||z_p||=||z_{n_i}||=||z_l||=1$.

\begin{figure*}[t]
\begin{center}
     \includegraphics[width=0.98\linewidth]{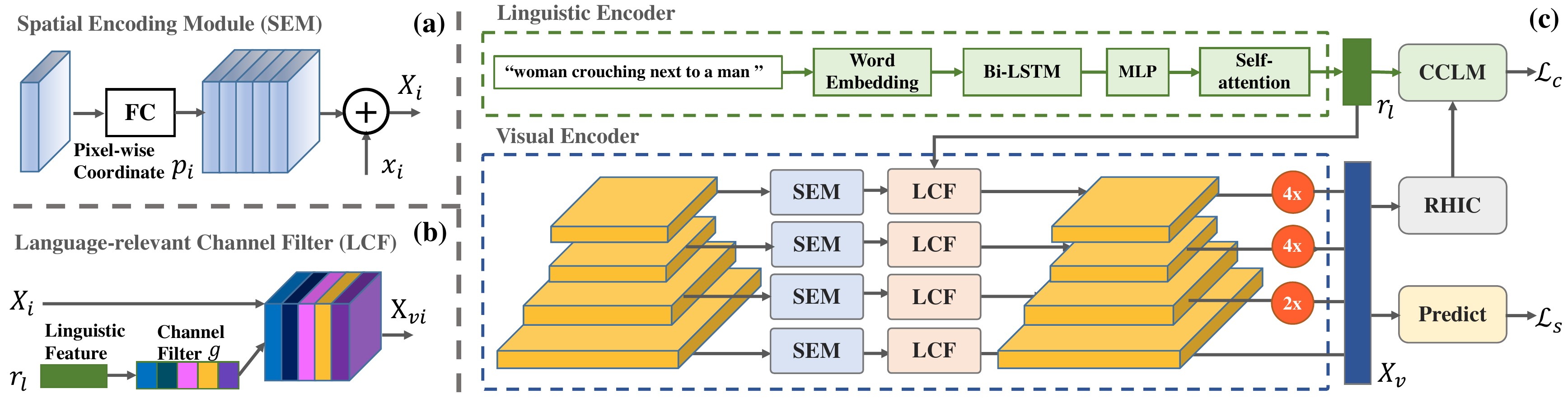}
\end{center}
\captionsetup{font=small}
\caption{\small
\textbf{Detailed architecture of CVLS pipeline.}
 Visual feature and linguistic feature are extracted by a FPN-like encoder and a bi-LSTM \cite{huang2015bidirectional} respectively \textbf{(c)}.
Language-relevant Channel Filter (\textbf{LCF}) (Sec.~\ref{section:LCF}) is integrated in visual encoder to exclude the over-simple visual-distinguishable feature.
Multi-scale output of the visual encoder is then concatenated as $X_v$ and fed into the Relative Hard Instance Construction (\textbf{RHIC}) (Sec.~\ref{section:LCF}) to aggregate instance-level visual features.
``\textbf{CCLM}" denotes the Cross-modal Contrastive Learning Module (Sec.~\ref{section:ObjCons}) as in Fig.~\ref{fig:pipeline}.
Detailed process of \textbf{(a)} Spatial Encoding Module (\textbf{SEM}) and \textbf{(b)} Language-relevant Channel Filter \textbf{(LCF)}.
}\label{fig:overall}
\end{figure*}

\begin{figure}[t]
\begin{center}
     \includegraphics[width=0.95\linewidth]{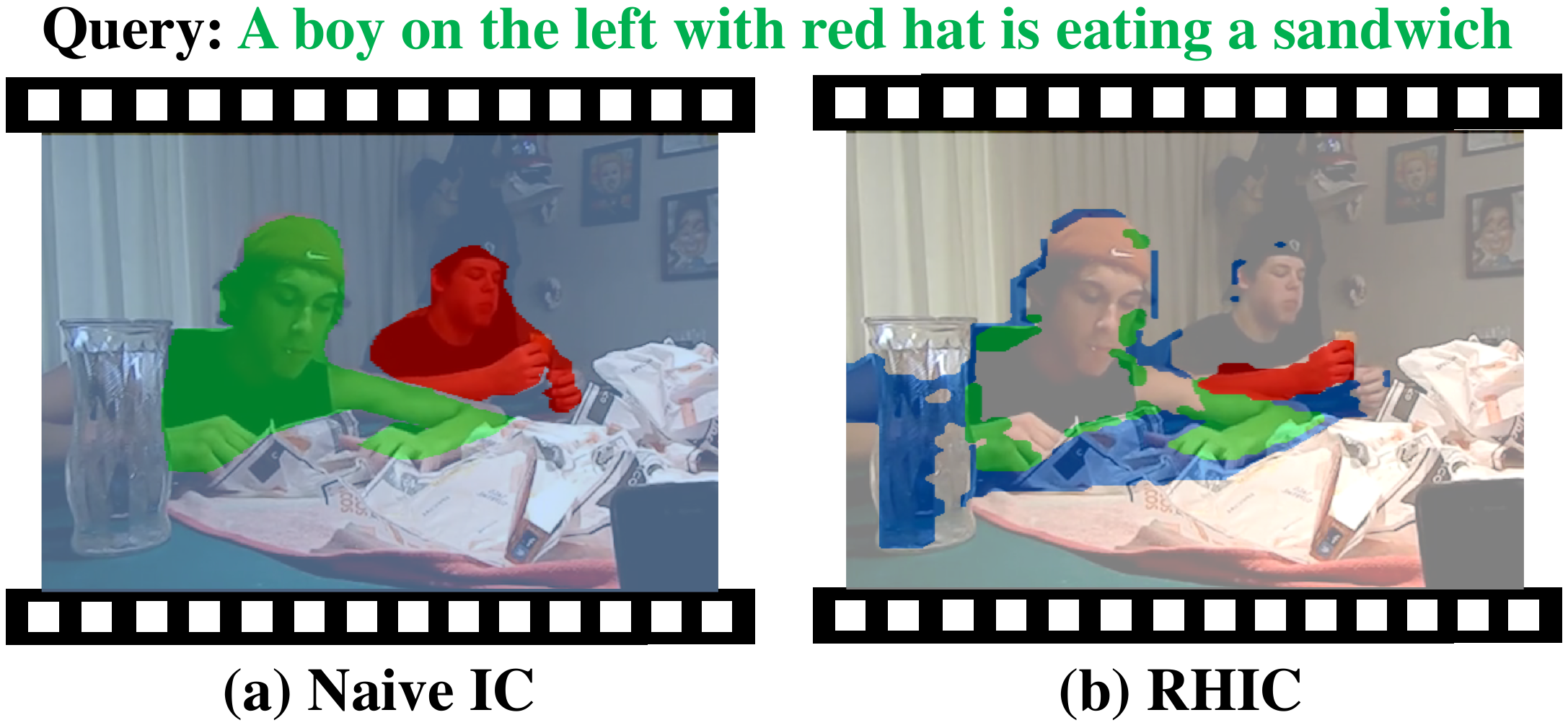}
\end{center}
\captionsetup{font=small}
\caption{\small
\textbf{Illustration of region-of-interests} while aggregating visual instance representation between Naive Instance Construction (Naive IC) and Relative Hard Instance Construction (RHIC).
Instead of pooling all the pixels, RHIC only considers relatively harder region of each instance, based on the confidence score of each pixel.
The positive referent and negative candidate are painted in green and red respectively.
Background region is also considered as a negative instance and is painted in blue in both figures.
}\label{fig:rhic}
\end{figure}

\subsubsection{ Instance-level Cross-Modal Contrast.}
With constructed instance-level features, the representation space is then shaped with the instance-level cross-modal contrastive objective $\mathcal{L}_c$, which optimizes the triple representations of the positive referent, the negative objects and the language description in the latent space:

\begin{equation}
\mathcal{L}_c = - \log \frac{\exp({z_p} \cdot z_l/\tau)}{\sum_{j} \exp({z_{n_j}} \cdot z_l/\tau) + \exp({z_p} \cdot z_l/\tau)},
\end{equation}
\noindent
where $\tau$ is a scalar temperature parameter, and a lower temperature value would increase the influence of harder negatives \cite{wu2018unsupervised,gunel2020supervised}.
The cross-modal contrastive objective regularizes the embedding space and force the network to explore explicit cross-modal discrimination among semantically similar objects.

\subsubsection{Overall Training Objective.}

During contrastive training, our method is supervised under both task-specific segmentation loss $\mathcal{L}_s$ and instance-level cross-modal contrastive loss $\mathcal{L}_c$ defined in previous section.
$\mathcal{L}_s$ optimizes the segmentation mask based on the
extracted multi-modal tensor:
\begin{equation}
    \mathcal{L}_s = - s \odot \log (\sigma(e)) - (1-s) \odot \log(1-\sigma(e)),
\end{equation}
where the $\cdot$ symbol denotes the inner (dot) product, $\odot$ is element-wise multiplication. And $e$ is the output of mask decoder and $s$ denotes the ground truth.
And overall training target of our method is a weighted average of $\mathcal{L}_s$ and $\mathcal{L}_c$:
\begin{equation}
    \mathcal{L} = \mathcal{L}_s + \lambda \mathcal{L}_c,
\end{equation}
\noindent
where $\lambda$ is a weighting hyperparameter, and $\lambda \in [0.1,1]$.

\subsection{Cross-Modal Hard Example Mining}\label{section:HM}
Though vanilla contrastive learning provides an explicit standpoint for modeling the fine-grained vision-language interaction, as we described in Sec.\ref{sec:introduction}, equally treat the visual-distinguishable regions and the linguistic-distinguishable objects (\textit{e.g.}, average pooling) might directly degrade the cross-modal contrastive objective to a single-modal feature classifier.
We thus propose two hard mining strategies tailored for cross-modal scenario, aiming to reduce the impact of excessive uninformative visual samples and enlarge the informative gains from both implicit and explicit aspects.

\subsubsection{Language-relevant Channel Filter (LCF). } \label{section:LCF}
As stated in~\cite{hu2018squeeze,chen2017sca,zeiler2014visualizing}, channel-wise features intrinsically encode the responses of different convolutional filters, and the visual concepts can be integral with different combination of channel responses.
Following this insight, we employ natural language as a filter to enhance language-relevant visual features and depress the language-irrelevant feature before contrastive training by applying dynamically conditioned weights on channels.
Specifically, we design a simple gating mechanism with a sigmoid activation as a filter on each channel.
With the linguistic feature $r_l$ and visual feature map $X$, the filter can be employed through:
\begin{equation}
    \begin{aligned}
        g &= \sigma(W_2\ \delta ({W_1}{r_l} + b_1)+b_2), \\
        X_v &= g \odot X,
    \end{aligned}
\end{equation}
where $\delta(\cdot)$ refers to the LeakyRELU operation \cite{xu2015empirical}. $W_i$ and $b_i$ are learnable matrix and bias.
Rather than enforcing the one-hot activation, we employ sigmod function $\sigma(\cdot)$ to normalize the excitation range to $(0,1)$, which enables a non-mutually-exclusive relationship among all channels. In this way, the impact of the visual-distinguishable feature declines, which implicitly forces the network to focus on linguistic-distinguishable content during the contrastive training.

\subsubsection{Relative Hard Instance Construction (RHIC). } \label{section:rhic}

An explicit hard mining strategy is introduced to further reduce the impact of uninformative visual regions. Dubbed Relative Hard Instance Construction (RHIC), it explicitly selects a portion of the entire object as \textit{hard} regions for visual object representation construction.
Specifically, during instance construction, only the misclassified pixels with high confidence scores are considered as \textit{hard} regions, which are sampled with a relative ratio (1:3) of hard to easy pixels.
Formally, given the model’s estimated probability $c$ predicted by segmentation decoder, we define misclassified degree $c_p$:

\begin{equation}
    \begin{aligned}
        c_p =
            \begin{cases}
                1-c   & \text{if $y=1$} \\
                c & \text{otherwise}.
            \end{cases}
    \end{aligned}
\end{equation}
In the above $y\!\in\!\{0,1\}$ specifies ground-truth label of each pixel.
The instance region in Equation \ref{equation:OT} can be updated by:
\begin{equation}
o^t = Top\_K(c_p) \odot o^t,
\end{equation}
where $Top\_K$ denotes the relative selecting process.
The visualization of the selected \textit{hard} regions by RHIC is shown in Fig.~\ref{fig:rhic}(b).
It is worth-noted that we employ a \textit{relative} selecting strategy here which introduces a flexible value of negative samples and $K$ corresponding to the total
pixel number of the object region and the prediction will be \textit{progressively} improved by \textit{online} selecting the relative-hard regions in each training step.
In practise, as occupation of different objects vary dramatically, we find a fixed threshold, \textit{i.e.}, a fixed number of $K$ is hard to tune. And we conduct additional ablations on relative ratio in Sec.\ref{sec:exp}.

\subsection{Overall Architecture}\label{section:overall}
In this section we instantiate the detailed architecture design as shown in Fig.~\ref{fig:overall}. Nevertheless, the idea of CVLS is general in the sense that basic encoder and decoder can be easily replaced with other stronger baselines.

\subsubsection{Linguistic Feature Encoder. } \label{section:LE}
Given a linguistic sentence $S = \{s_i\}_{N_l}$ with $N_l$ words,
linguistic feature is encoded with a bi-LSTM \cite{huang2015bidirectional} module followed by a self-guided attention fusion.
Specifically, linguistic feature $r_l \in \mathbb{R}^{C_v}$ can be calculated from:
\begin{equation}
r_l = W_l(\sum_{i=1}^{N_l}{\alpha_{i}h_i}) + b_l,
\end{equation}
where $h_i$ is the corresponding hidden state of $i$-th word, $\alpha$ is the word-level attention weights derived from: $\alpha_i = softmax(fc(h_i))$. $W_l$ and $b_l$ are learnable parameters. $r_l$ is projected to the feature visual feature space with $C_v$ dimensions.
he self-guided attention fusion introduces a flexible way for highlighting keywords in a sentence and reduce the negative impact caused by sentence truncation or padding.

\subsubsection{Visual Feature Encoder. }\label{section:VE}
As illustrated in Fig.~\ref{fig:overall}(c), our network takes a video clip $\mathcal{V}=\{f_i\}_{T}$ as input.
For each frame, the multi-level visual features $\{x_i\}$ are extracted with a CNN backbone (\textit{e.g.}, ResNet-50 \cite{he2016deep}) and respectively fused with an 8-D spatial coordinate feature $p_i \in \mathbb{R}^{H_i \times W_i \times 8}$ ($i \in \{2,3,4,5\}$), where $H_i$ and $W_i$ are the height and width of the visual features.
We collect the spatial augmenting operations as a Spatial Encoding Module (SEM) and illustrates it in Fig.~\ref{fig:overall}(a). It is designed for remedying the weak positional sensation of spatial-agnostic CNN-extracted visual feature.
Formally, the spatial enhanced visual feature $X_i \in \mathbb{R}^{H_i \times W_i \times C_v^i}$ at level $i$ can be calculated from:
\begin{equation}
X_i = x_i + W_p(p_i),
\end{equation}
where $W_p$ is a learnable matrix and $C_v^i$ is the channel dimension.
$\{X_i\}$ are then enhanced with multi-scale information by merging with features from a top-down pathway via simple upsampling and addition, like in \cite{kirillov2019panoptic}.
Multi-level features are then concatenated to obtain final feature map, which would be used for task-specific mask prediction and contrastive learning ($X_v$ in Sec.~\ref{section:ObjCons}).

\setlength{\tabcolsep}{5pt}
\begin{table*}
\small
\begin{center}
\resizebox{1.\textwidth}{!}{
		\setlength\tabcolsep{8pt}
		\renewcommand\arraystretch{1.0}
\begin{tabular}{l|c|ccccc|c|cc}
\hline
\multirow{2}{*}{\textbf{Methods}} & \multirow{2}{*}{\textbf{Venue}}      & \multicolumn{5}{c|}{\textbf{Overlap}}                                                                          & \textbf{mAP} & \multicolumn{2}{|c}{\textbf{IoU}} \\
 & & P@0.5 & P@0.6 & P@0.7 & P@0.8 & P@0.9 & 0.5:0.95 & Overall & Mean \\
\shline
\cite{hu2016segmentation} & ECCV16 & 34.8 & 23.6 & 13.3 & 3.3  & 0.1 & 13.2 & 47.4 & 35.0 \\
\cite{li2017tracking} & CVPR17 & 38.7 & 29.0 & 17.5 & 6.6  & 0.1 & 16.3 & 51.5 & 35.4 \\
\cite{gavrilyuk2018actor} & CVPR18 & 53.8 & 43.7 & 31.8 & 17.1 & 2.1 & 26.9 & 57.4 & 48.1 \\
\cite{wang2019asymmetric} & ICCV19 & 55.7 & 45.9 & 31.9 & 16.0 & 2.0 & 27.4 & 60.1 & 49.0 \\
\cite{mcintosh2020visual} & CVPR20 & 52.6 & 45.0 & 34.5 & 20.7 & 3.6 & 30.3 & 56.8 & 46.0 \\
\cite{ningpolar} & IJCAI20 & 63.4 & 57.9 & 48.3 & 32.2 & 8.3 & 38.8 & 66.1 & 52.9 \\
\cite{wang2020context} & AAAI20 & 60.7 & 52.5 & 40.5 & 23.5 & 4.5 & 33.3 & 62.3 & 53.1 \\
\cite{hui2021collaborative} & CVPR21 & \textbf{65.4} & \textbf{58.9} & 49.7 & 33.3 & 9.1 & 39.9 & 66.2 & 56.1 \\
\hline
\textbf{Ours} & - & 64.2 & 58.7 & \textbf{50.5} & \textbf{35.8} & \textbf{11.6} & \textbf{40.8} & \textbf{68.3} & \textbf{57.1} \\
\hline
\end{tabular}
}
\end{center}
\captionsetup{font=small}
\caption{\small
Comparison with the state-of-the-art methods on the A2D Sentences \texttt{valid} using \textit{IoU} and \textit{Precision@K} as metrics.
}
\label{table:a2dsota}
\end{table*}

\setlength{\tabcolsep}{5pt}
\begin{table*}
\small
\begin{center}
\resizebox{1.\textwidth}{!}{
		\setlength\tabcolsep{8pt}
		\renewcommand\arraystretch{1.0}
\begin{tabular}{l|c|ccccc|c|cc}
\hline
\multirow{2}{*}{\textbf{Methods}} &  \multirow{2}{*}{\textbf{Venue}}      & \multicolumn{5}{c|}{\textbf{Overlap}}                                                                          & \textbf{mAP} & \multicolumn{2}{|c}{\textbf{IoU}} \\
 & & P@0.5 & P@0.6 & P@0.7 & P@0.8 & P@0.9 & 0.5:0.95 & Overall & Mean \\
\shline
\cite{hu2016segmentation} & ECCV16          & 63.3 & 35.0 & 8.5 & 0.2 & 0.0 & 17.8 & 54.6 & 52.8 \\
\cite{li2017tracking} & CVPR17             & 57.8 & 33.5 & 10.3 & 0.6 & 0.0 & 17.3 & 52.9 & 49.1 \\
\cite{gavrilyuk2018actor} & CVPR18   & 71.2 & 51.8 & 26.4 & 3.0 & 0.0 & 26.7 & 55.5 & 57.0 \\
\cite{wang2019asymmetric} & ICCV19       & 75.6 & 56.4 & 28.7 & 3.4 & 0.0 & 28.9 & 57.6 & 58.4 \\
\cite{mcintosh2020visual} & CVPR20   & 67.7 & 51.3 & 28.3 & 5.1 & 0.0 & 26.1 & 53.5 & 55.0 \\
\cite{ningpolar} & IJCAI20                & 69.1 & 57.2 & 31.9 & 6.0 & 0.1 & 29.4 & - & - \\
\cite{wang2020context} & AAAI20          & 74.2 & 58.7 & 31.6 & 4.7 & 0.0 & 30.1 & 55.4 & 57.6 \\
\cite{hui2021collaborative} & CVPR21          & 78.3 & 63.9 & 37.8 & 7.6 & 0.0 & 33.5 & 59.8 & 60.4 \\
\hline
\textbf{Ours} & - & \textbf{86.0} & \textbf{72.7} & \textbf{44.8} & \textbf{9.7} & \textbf{0.1} & \textbf{39.2} & \textbf{66.1} & \textbf{64.3} \\
\hline
\end{tabular}
}
\end{center}
\captionsetup{font=small}
\caption{\small
Comparison with the state-of-the-art methods on J-HMDB Sentences \texttt{test} using \textit{IoU} and \textit{Precision@K} as metrics.
}
\label{table:jhmdbsota}
\end{table*}

\begin{table}
\small
\begin{center}
\resizebox{0.46\textwidth}{!}{
		\setlength\tabcolsep{8pt}
		\renewcommand\arraystretch{1.1}
\begin{tabular}{ccc|c|cc}
\hline
\multicolumn{3}{c|}{\textbf{Modules}} & \textbf{mAP} & \multicolumn{2}{|c}{\textbf{IoU}} \\
CCL & RHIC & LCF & 0.5:0.95 & Overall & Mean \\
\shline
 &  &  & 26.6 & 56.9 & 47.1 \\
$\surd$ &  &  & 34.4 & 64.2 & 52.7 \\
$\surd$ & $\surd$ &  & 36.2 & 65.6 & 53.3 \\
$\surd$ &  & $\surd$ & 39.7 & 67.5 & 56.8 \\
$\surd$ & $\surd$ & $\surd$ & \textbf{40.8} & \textbf{68.3} & \textbf{57.1} \\
\hline
\end{tabular}
}
\end{center}
\captionsetup{font=small}
\caption{\small
\textbf{Component analysis} on A2D Sentences \texttt{valid}.
}
\label{table:ablation}
\end{table}

\begin{figure*}[t]
\begin{center}
     \includegraphics[width=.86\linewidth]{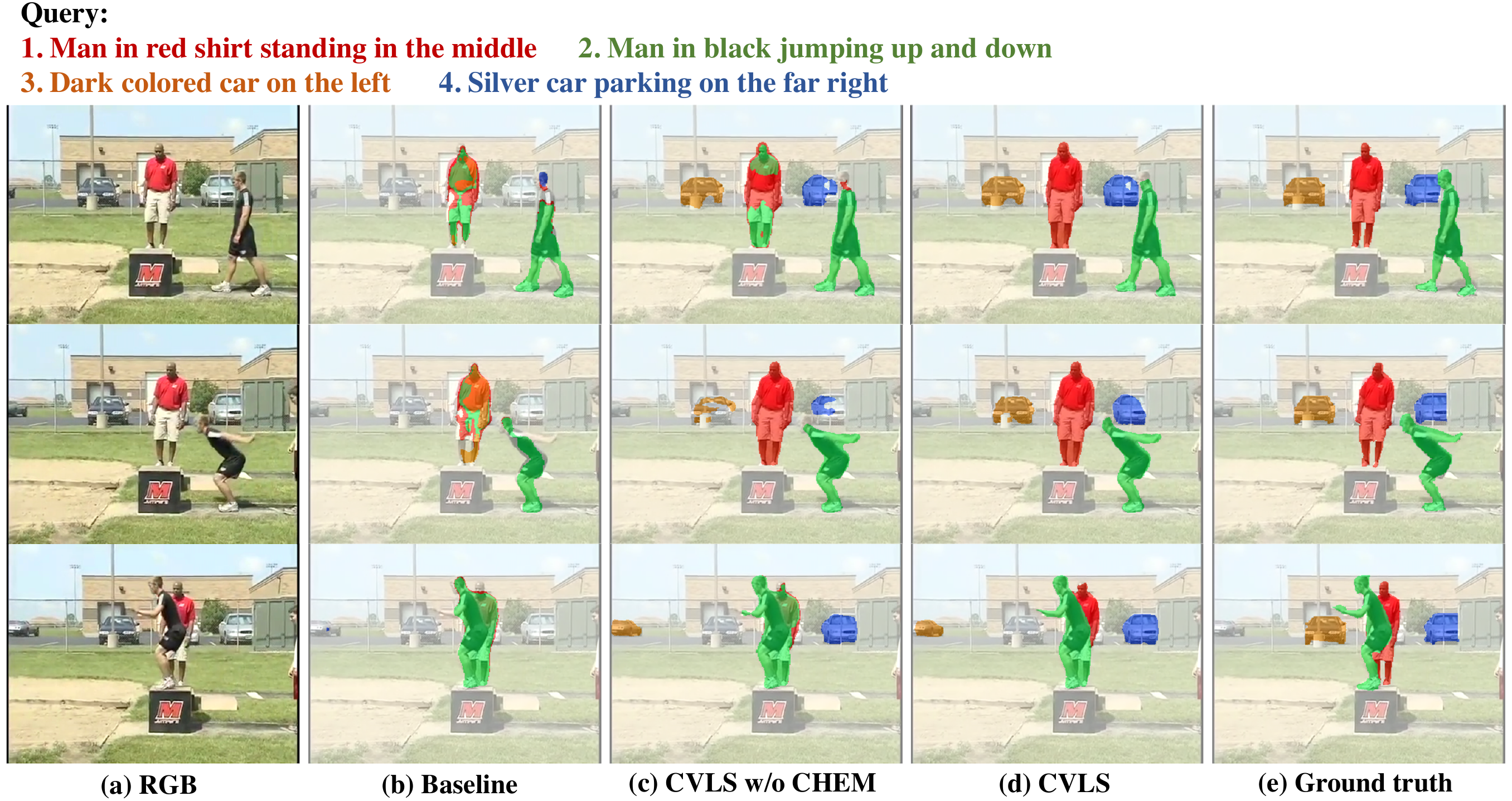}
\end{center}
\captionsetup{font=small}
\caption{\small
      \textbf{Qualitative results} of VLS on A2D Sentences.
      Predicted result is represented by the same color with the corresponding sentence.
      Cross-modal Hard Example Mining is abbreviated as ``CHEM".
     (a) Original image. (b)-(d) are predicted results by:
     (b) Baseline model, \textit{i.e.}, implicit fusion (row 1 in Table \ref{table:ablation});
     (c) CVLS \textit{w/o} CHEM (row 2 in Table \ref{table:ablation});
     (d) Full CVLS (row 5 in Table \ref{table:ablation}).
     (e) Ground truth.
      }\label{fig:qualiti}
\end{figure*}

\section{Experiments}\label{sec:exp}

\subsection{Datasets and Evaluation Criteria}
All experiments in this paper are conducted on two extended datasets: \textbf{A2D Sentences} and \textbf{J-HMDB Sentences}. These datasets are released in \cite{gavrilyuk2018actor} by additionally providing natural descriptions on original A2D \cite{xu2015can} and J-HMDB \cite{jhuang2013towards} respectively.
\noindent\textbf{A2D Sentences} contains 3782 videos in total with 8 action classes performed by 7 actor classes. Each video in A2D has 3 to 5 frames annotated with pixel-level actor-action segmentation masks. Besides, it contains 6,655 sentences corresponding to actors and their actions. Following the settings in \cite{wang2019asymmetric}, we split the whole dataset into 3017 training videos, 737 testing videos, and 28 unlabeled videos.
\noindent\textbf{J-HMDB Sentences} contains 928 short videos with 928 corresponding sentences describing 21 different action classes. Pixel-wise 2D articulated human puppet masks are provided for evaluating segmentation performance.

We evaluate our proposed method with the criteria of Intersection-over-Union (IoU) and precision. Overall IoU is the ratio of the total intersection area divided by the total union area over testing samples. The mean IoU is the averaged IoU over all samples, which may not be affected by the size of samples. We also measure precision@K which considers the percentage of testing samples whose IoU scores are higher than threshold K and calculate mean average precision over 0.50:0.05:0.95 \cite{gavrilyuk2018actor}.

\subsection{Implementation Details}
We adopt ResNet50 \cite{he2016deep} model pre-trained on ImageNet \cite{deng2009imagenet} as visual backbone and use the output of Res2, Res3, Res4 and Res5 for multi-level feature construction. And we instantiate mask decoder with three stacked $3\times3$ convolution layers for decoding followed by one $1\times1$ convolutional layer for outputting the final segmentation mask.
A bi-LSTM~\cite{huang2015bidirectional} module is utilized as text encoder. All input frames are resized to $320 \times 320$.
Following the settings in \cite{gavrilyuk2018actor}, the maximum length of sentences is set to 20 and the dimension of word vector is 1000. The word vectors are initialized with one-hot vectors without any pre-trained weights applied. The hidden states of bi-LSTM \cite{huang2015bidirectional} are encoded as sentence features with a dimension of 2000.
%
%
Training is done with Adam optimizer \cite{kingma2014adam} with an initial learning rate of $0.0002$. We employ a scheduler that waits for $2$ epochs after loss stagnation to reduce the learning rate by a factor of $10$ and the batch size is set to $8$ by default.
During inference, following the setting in \cite{wang2020context}, we take a pixel as foreground when its value is higher than $\beta$ of the maximum value in probability map. And $\beta$ is set to $0.8$ in our implementation.

\subsection{Comparison with State-of-the-Art Methods}
Following the standard settings, we compare our CVLS with other state-of-the-art video-language segmentation models published in recent years on two datasets: A2D Sentences and J-HMDB Sentences. The comparison results are demonstrated in Table \ref{table:a2dsota} and Table \ref{table:jhmdbsota}.
First on A2D Sentences, baseline methods \cite{hu2016segmentation,li2017tracking} are pre-trained on ReferIt dataset \cite{kazemzadeh2014referitgame} and then fine-tuned on A2D sentences. Other methods including ours are trained on A2D Sentences exclusively.
As shown in Table \ref{table:a2dsota}, we bring 1.2$\%$ improvement on Mean IoU and 2.1$\%$ on Overall IoU over SOTA respectively, which directly proves the effectiveness of our method.
The state-of-the-art performance has been achieved on most metrics especially at higher IoU thresholds, \textit{i.e.}, $P@0.8$ and $P@0.9$.
On $P@0.8$, our method outperforms the SOTA by a margin of 2.5$\%$. On $P@0.9$, 2.5$\%$ absolute improvement is achieved with a 27$\%$ relative improvement.
Then, we conduct experiments on the whole J-HMDB Sentences to evaluate the generalization ability of our CVLS. As a default setting, the best model trained on A2D sentences is employed without any additional fine-tuning.
With finer cross-modal interaction, our CVLS significantly outperforms previous state-of-the-art methods on all metrics considered.

\subsection{Ablation Study}
\subsubsection{Component Analysis. }
\noindent\textit{Settings:}
We conduct extensive component analysis to verify the effectiveness and reveal the internal mechanism of each component in CVLS. We summarize ablation results of each proposed module in Table \ref{table:ablation}.
Baseline, in line1, ignores the fine-grained cross-modal interaction and directly predicts segmentation mask with implicit concatenation-convolution~\cite{shi2018key,hu2016segmentation} fusion.
In other lines, the explicitly cross-modal interaction is involved and the network with different components is supervised under a combination of cross-modal contrastive loss and segmentation loss.

\noindent\textit{Observations:}
All ablations are conducted on the validation set of A2D Sentences.
We could obtain the following observations:
\textbf{(1)} Due to the redundant uninformative visual-distinguishable feature, the vanilla CCL merely brings limited improvement compared with baseline.
\textbf{(2)} Both of the hard mining strategies can significantly improve the vanilla CCL, which validates the effects of the design.
\textbf{(3)} Conclusively, these results confirm the merits of the explicitly cross-modal interaction formulation again.

\begin{figure}[t]
\begin{center}
     \includegraphics[width=.98\linewidth]{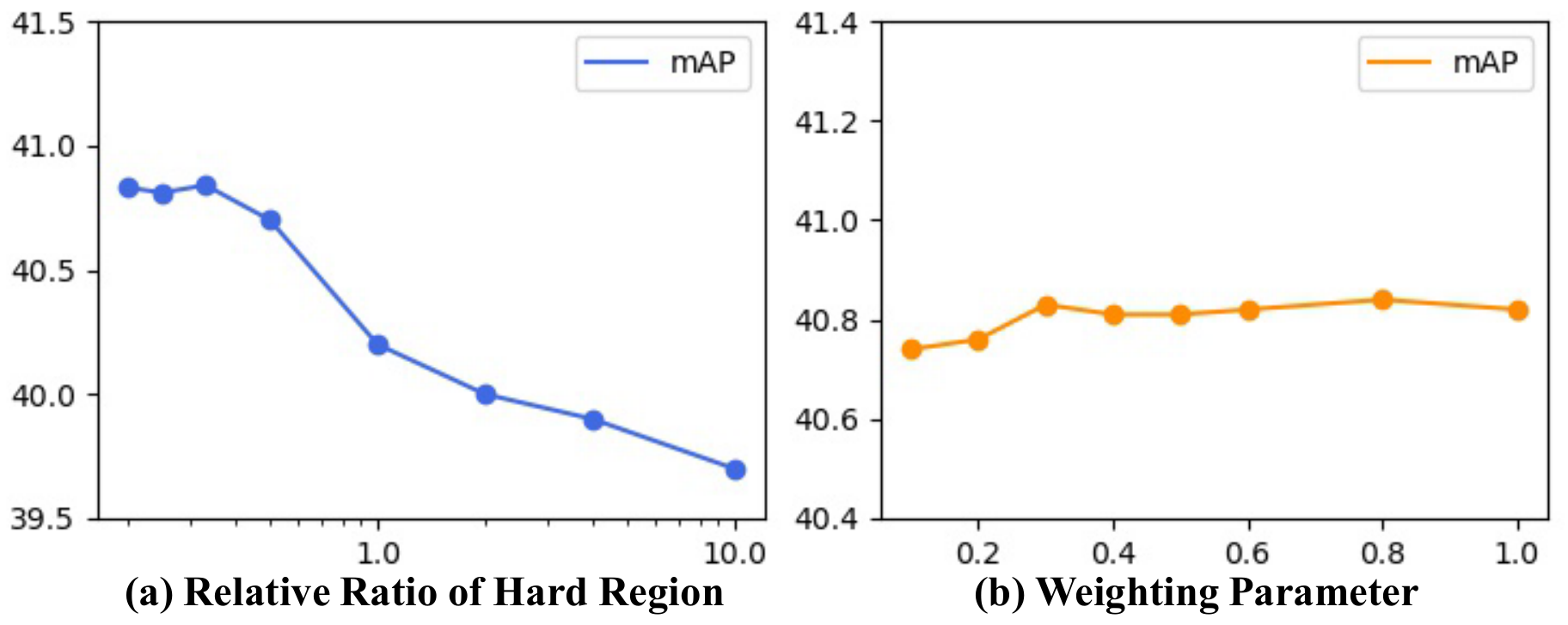}
\end{center}
\captionsetup{font=small}
\caption{\small
\textbf{Variation of model performance} with different selecting ratios in RHIC (Sec.~\ref{section:rhic}) and different objective weighting $\lambda$.
}\label{fig:sensitive}
\end{figure}

\subsubsection{Relative ratio of \textit{hard} samples in RHIC. }
We supply an additional experiment on ranges of relative ratio of RHIC in Fig.\ref{fig:sensitive}~(a). We modify relative ratio of hard to easy pixels from 1:2 to 1:5, and final performance of CVLS is stable ($\pm$0.15$\%$ in terms of mAP) but extra epochs are needed for convergence. While the ratio is ranged from 1:1 to 4:1, final performance progressively drops $\sim$0.9$\%$, as RHIC is progressively losing efficacy which causes model degradation. We empirically set the ratio to 1:3 for balancing the performance and convergence speed.

\subsubsection{Parameter sensitivity analysis of $\lambda$. } As we employ a joint loss of segmentation term and contrastive term, we further evaluate different values of the weighting parameter $\lambda$ (Sec.~\ref{section:ObjCons}) and summarize the results
in Fig.~\ref{fig:sensitive}~(b). In practice, we find the proposed approach is robust to parameter variation and the performance is slightly jittered among different settings. We finally set $\lambda$ to 0.8 empirically.

\subsection{Qualitative Analysis}
In this section, we investigate the internal mechanism in CVLS by analyzing qualitative results.
The comparisons between the full model and alternative structures are illustrated in Fig.~\ref{fig:qualiti}.
Compared with the implicit fusion baseline (Fig.~\ref{fig:qualiti}(b)), even without hard mining strategies, the vanilla cross-modal contrastive learning pipeline could still capture finer cross-modal interaction and shows reasonable segmentation results.
Due to the model degradation caused by extra visual-distinguishable feature, the model still messes up the semantically similar region, \textit{i.e.}, two guys with red and green masks respectively (the third line in Fig.~\ref{fig:qualiti}(c)). While in Fig.~\ref{fig:qualiti}(d), the problem is greatly alleviated.
Conclusively, these visualized results reconfirm the effectiveness of CVLS and the hard mining strategies which facilitates the entire method in both implicit (LCF) and explicit (RHIC) aspects.

\section{Conclusion}
In this paper, we propose a highly effective pipeline with a novel cross-modal contrastive learning objective
for the first attempt of introducing the explicit cross-modal interaction into video-language segmentation field.
To prevent the model from degradation to single-modal, we further employ two cross-modal hard mining strategies to exclude the over-simple visual-distinguishable feature from both implicit and explicit aspects, \textit{i.e.}, Language-relevant Channel Filter and Relative Hard Instance Construction.
Evaluations on commonly used benchmarks demonstrate that CVLS surpasses all the state-of-the-art methods by large margins.

\bibliography{aaai22}

\end{document}